\pdfoutput=1

\documentclass[11pt]{article}

\usepackage[final]{acl}
\usepackage{times}
\usepackage{latexsym}

\usepackage[T1]{fontenc}

\usepackage[utf8]{inputenc}

\usepackage{microtype}

\usepackage{inconsolata}

\usepackage{graphicx}

\usepackage{todonotes}

\usepackage{microtype}

\usepackage{wrapfig}
\usepackage{array}

\usepackage{todonotes}
\usepackage{amsmath}
\usepackage{comment}
\usepackage{dialogue}
\usepackage{mdframed}
\usepackage{environ}
\usepackage{algorithm}

\usepackage{algpseudocode}
\usepackage{amssymb} 
\usepackage{xcolor}
\usepackage[italian,english]{babel}

\usepackage{multirow}
\usepackage{booktabs} 
\usepackage{rotating}
\usepackage{enumitem}
\usepackage{tabularray}
\usepackage{tikz}

\newcommand{\dialoguetable}[4]{
    \begin{table*}[h!]
        \small
        \centering
        \begin{tikzpicture}
        \node (table) [inner sep=0pt] {
        \begin{tabular}{p{0.82\linewidth}}\
        \\
        \textbf{\textsc{Dialogue #1}} \\
        #2 \\
        \\
        \end{tabular}
        };
        \draw [rounded corners=.5em, very thick] (table.north west) rectangle (table.south east);
        \end{tikzpicture}
        \caption{#3}
        \label{tab:#4}
    \end{table*}
}

\newcommand{\humanevalguidelines}[4]{
    \begin{table}[h!]
        \small
        \centering
        \begin{tikzpicture}
        \node (table) [inner sep=0pt] {
        \begin{tabular}{p{0.82\linewidth}}\
        \\
        \textbf{\textsc{Human Evaluation Guidelines #1}} \\
        #2 \\
        \\
        \end{tabular}
        };
        \draw [rounded corners=.5em, very thick] (table.north west) rectangle (table.south east);
        \end{tikzpicture}
        \caption{#3}
        \label{tab:#4}
    \end{table}
}

\newcommand{\prompttable}[4]{
    \begin{table}[h!]
        \small
        \centering
        \begin{tikzpicture}
        \node (table) [inner sep=0pt] {
        \begin{tabular}{p{0.82\linewidth}}\
        \\
        \textbf{\textsc{Prompt #1}} \\
        #2 \\
        \\
        \end{tabular}
        };
        \draw [rounded corners=.5em, very thick] (table.north west) rectangle (table.south east);
        \end{tikzpicture}
        \caption{#3}
        \label{tab:#4}
    \end{table}
}

\title{When Harry Meets Superman:\\ 
The Role of The Interlocutor in Persona-Based Dialogue Generation
}

\author{
 Daniela Occhipinti$^{1,2}$,
 Marco Guerini$^1$,
 Malvina Nissim$^3$
 \\
 $^1$Fondazione Bruno Kessler, Via Sommarive 18, Povo, Trento, Italy\\
 $^2$University of Trento, Italy\\
 $^3$University of Groningen, Netherlands
 \\
 \texttt{docchipinti@fbk.eu, guerini@fbk.eu, m.nissim@rug.nl}
}

\begin{document}
\maketitle
\begin{abstract}

Endowing dialogue agents with persona information has proven to significantly improve the consistency and diversity of their generations. While much focus has been placed on aligning dialogues with provided personas, the adaptation to the \textit{interlocutor}’s profile remains largely underexplored. In this work, we investigate three key aspects: (1) a model's ability to align responses with both the provided persona and the interlocutor’s; (2) its robustness when dealing with familiar versus unfamiliar interlocutors and topics, and (3) the impact of additional fine-tuning on specific persona-based dialogues. We evaluate dialogues generated with diverse speaker pairings and topics, framing the evaluation as an author identification task and employing both LLM-as-a-judge and human evaluations. By systematically masking or disclosing information about the interlocutor, we assess its impact on dialogue generation. Results show that access to the interlocutor's persona improves the recognition of the target speaker, while masking it does the opposite. Although models generalise well across topics, they struggle with unfamiliar interlocutors. Finally, we found that in zero-shot settings, LLMs often copy biographical details, facilitating identification but trivialising the task.

\end{abstract}

\section{Introduction}

\begin{figure}[!t]
\centering
\includegraphics[width=1\columnwidth]{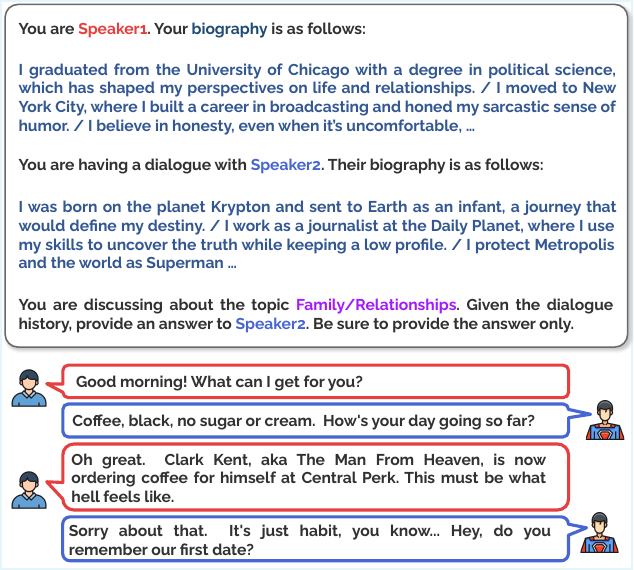}
 \caption{Example of a generated dialogue between Harry and Superman. The upper section displays the prompt, while the lower the beginning of the generated conversation.}
 \label{fig:dialogue_esempio}
 \end{figure}

In recent years, large language models (LLMs) have proven effective in generating coherent and contextually appropriate responses. 
One of their applications is in 
role-play systems \cite{chen2024persona, tseng2024talespersonallmssurvey}, where they 
simulate conversations by adapting to a specified \textit{target} persona in a given prompt (e.g., "You are \textit{a young woman passionate about cinema"}). Indeed, research has amply focused on developing and studying models that adapt their speech patterns, style, and content to align to a target persona they have to represent \cite{lu2024large, wang-etal-2024-incharacter}. 

One aspect that has received much less attention instead, is the ability of such persona-based agents to adapt to the other speaker they are engaging with (i.e., the \textit{interlocutor}). Indeed, the impact on the model's responses of the interlocutor's persona, which plays a relevant role in yielding a more coherent and appropriate conversation, has been explored only in a few seminal works \citep{liu-etal-2020-impress, gu2021partner, xu2022cosplay, lu-etal-2022-partner}. While these works observe some degree of adjustment to the other speaker, the actual adaptation to interlocutor and topic variations is not systematically evaluated, especially regarding its effect on target speaker identification.

\begin{figure*}[h]
    \centering
    \includegraphics[width=1\textwidth]{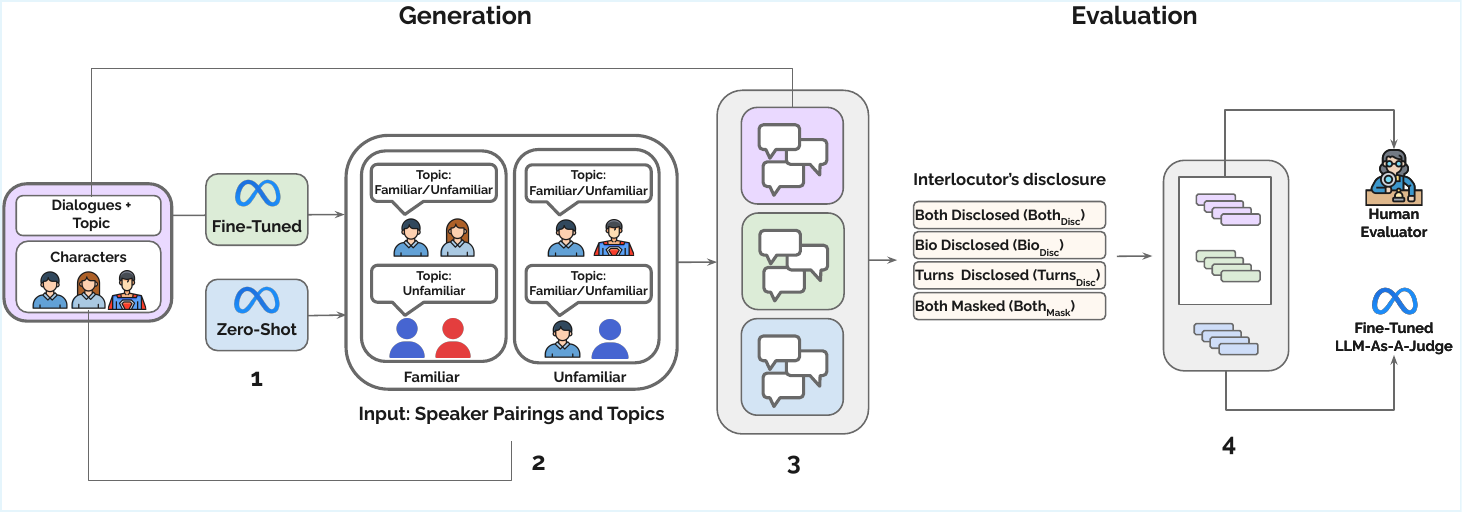}
  \caption{
Pipeline: 
(i) Fine-tune Llama 3.1 8B Instruct on PRODIGy dialogues and speaker pairings.
(ii) Create speaker pairings using PRODIGy and newly created Non-PRODIGy characters.
(iii) Generate dialogues using both the fine-tuned model and a zero-shot setting providing the created pairings as input.
(iv) Evaluate dialogues using LLM-as-a-Judge (for all model outputs and gold dialogues) and human reviewers (for fine-tuned model outputs and gold dialogues only), systematically masking or revealing interlocutor information.}
    \label{fig:pipeline}
\end{figure*}

In the present research, we explore this aspect by addressing the following research questions:

\begin{description}[leftmargin=*]
\itemsep0em

\item[Q1] Does the model adapt more efficiently to the target speaker's persona, or to the interlocutor's persona and dialogue turns?

\item[Q2] To which aspects of the dialogue does the model adapt more effectively? The aspects we consider are the following:
\begin{itemize} [leftmargin=*]
    \item \textit{unfamiliar interlocutors} who fall outside the typical context of the target speakers (e.g., Harry from the movie \textit{When Harry Met Sally} talking with Superman, as in Figure~\ref{fig:dialogue_esempio});
    \item \textit{unfamiliar topics} to the target speaker (e.g., Harry from the movie \textit{When Harry Met Sally} having a conversation about war).
\end{itemize}

\item[Q3] Is additional fine-tuning on specific persona-based dialogues beneficial to the models' performance in preserving and adapting persona features?

\end{description}

\noindent To answer these questions, we generate dialogues with various speaker pairings and topics using both fine-tuned and zero-shot settings. We frame the task as author identification \cite{stamatatos2014overview}, where given a dialogue and a pool of possible personas, an evaluator has to assign the correct persona to the target speaker. We introduce a novel evaluation paradigm specifically tailored to assess the influence of interlocutors and topics on dialogue generation and speaker identification. In this framework (Figure~\ref{fig:pipeline}), we conduct extensive evaluations with human reviewers and an LLM-as-a-judge.

By systematically masking or revealing different aspects of the interlocutor's information, we measure how efficiently the LLM leverages the target speaker's information and adapts its responses to the interlocutor. 
The key idea is that if masking interlocutor information makes it harder to identify the target speaker’s biography, it is likely that the model has 
appropriately adapted its responses to the interlocutor's information, while still staying coherent to the target persona. For instance, if the target speaker's persona is \textit{a young woman passionate about cinema} and the interlocutor's is \textit{her little daughter}, the model adjusts its language to fit a mother-child interaction rather than focusing on cinema. Masking the interlocutor’s biography makes identifying the target speaker harder, as the conversation may not explicitly reference cinema. 

Our framework, revolving around masking and revealing interlocutor information during evaluation, allows us to assess more directly the impact of the interlocutor on both dialogue generation and target speaker identification. Our findings show that models adapt easily to any topic given to them, while they struggle more to adapt to unfamiliar interlocutors, highlighting the importance of 
\textit{who we are speaking with} over \textit{what we are speaking about}. Moreover, while zero-shot models can adapt to both target and interlocutor speakers, they tend to capture only surface-level persona traits: we demonstrate that fine-tuning significantly improves their ability to grasp deeper persona characteristics.

\section{Related Work}

\paragraph{Role-Play Agents}
Despite their impressive abilities, LLMs often fall short in capturing the deep characteristics of human traits and interactions \cite{shanahan2023role}. To address this limitation, Role-Play LLMs have been developed to simulate assigned personas, capturing their distinctive traits. A key research line is the \textit{character persona}, where models are provided with well-known fictional or real-world figures \cite{chen2024persona}. 
Recent research enhanced the model’s ability to consistently emulate specific personas by employing techniques such as memory reconstruction, long-term memory integration, and fine-grained persona construction \cite{shao-etal-2023-character, park-et-al2023, wang-etal-2024-rolellm, lu2024large}. 
However, while significant progress has been made in simulating individual characters, few studies have investigated how interlocutor information influences response generation \cite{xu2022cosplay, lu-etal-2022-partner, liu-etal-2020-impress, zhang-etal-2018-personalizing}. None of these studies has addressed how models adapt their responses across diverse speaker pairings and topics and do not directly evaluate how interlocutor information impacts dialogue generation by assessing its role in target speaker identification.

\paragraph{Role-Play Agents: Evaluation}
Evaluation for role-play models focuses on how well they replicate given characters, considering both superficial traits (linguistic style, knowledge) and deeper ones (identity, relationships, experiences) \cite{wang-etal-2024-rolellm, li2023chatharuhirevivinganimecharacter, zhou-etal-2024-characterglm, yu-etal-2024-neeko}. Early evaluation methods predominantly relied on overlap-based metrics such as ROUGE \cite{lin-2004-rouge}, BLEU \cite{papineni-etal-2002-bleu}, and METEOR \cite{banerjee-lavie-2005-meteor}, which compare generated text against reference responses. However, these metrics primarily capture surface-level elements and fail to assess deeper character traits. More recently, evaluator LLMs have been employed to score responses, evaluate quality or perform pairwise comparisons based on reference answers. For instance, \citet{wang-etal-2024-incharacter} used GPT-3.5 to evaluate role-playing agents across multiple dimensions, including memorization, values, hallucination, and stability. Similarly, \citet{lu2024large} and \citet{wu-etal-2024-role} employed GPT-4 \cite{openai2023gpt4} to assess role consistency, knowledge accuracy, and alignment with internal thought processes. 
Beyond automated methods, human evaluation plays is crucial to assess both the superficial and deeper aspects. However, it demands significant effort, lacks reproducibility, and is less scalable than automated methods.

\section{Experimental Design}

In this section, we describe the profile-based dialogue dataset we used and the methodology for assigning topics to each dialogue. Next, we introduce the models and the fine-tuning configuration we used for dialogue generation. Finally, we detail both automatic and human evaluation strategies. 

\subsection{Dataset}\label{dataset}

We employ the PRODIGy dataset \cite{occhipinti-etal-2024-prodigy}, comprising dialogues and characters from movies. The dataset includes comprehensive speaker profiles, including communication styles, biographies, personalities, and gender information. Each dialogue involves two speakers, with at least one speaker annotated with a profile.

We selected 5,660 dialogues where both speakers have profile annotations. This subset allows us to explore the influence of both target speaker and interlocutor's information on dialogue generation, with a focus on assessing the interlocutor's role. Among the various profile dimensions provided in PRODIGy, we focus on character biographies, as they offer rich semantic content, making them more relevant to our analysis.

By employing Llama 3.1 8B Instruct \cite{dubey2024llama}, we annotated dialogues with topic labels to guide generation and evaluate the models' ability to generalise across different themes (see Appendix~\ref{topic-annotation} for details). The dataset was then split into training, validation, and test sets (80:10:10), ensuring the same speaker pairs did not appear across different sets, though individual speakers might recur.

\subsection{Models and Generations}

We conducted experiments with Llama 3.1 8B Instruct \cite{dubey2024llama} in two settings: fine-tuning with PRODIGy and zero-shot prompting (see Appendix \ref{model-generation-appendix} for further details). Focusing on a single model allows us to attribute any observed differences to our fine-tuning methodology rather than to differences in model architecture or size, which was a key consideration of this study.  Notably, the same experiments can be reproduced under identical controlled settings using different model families, enabling broader investigations into persona-based generation and its generalisability across architectures.  We also excluded few-shot prompting, as the additional contextual cues it provides could obscure the model’s intrinsic ability to adapt to target and interlocutor personas. Furthermore, LLMs are known to be sensitive to the order of few-shot examples, and the increased diversity of the experimental conditions would have made consistent prompting infeasible, introducing substantial noise in the evaluation.

Since PRODIGy dialogues and characters are derived from movies and screenplays that are available online, there might be data contamination effects. To control for these effects, we designed experiments generating dialogues across diverse combinations of character pairings and topics, introducing incremental randomness to evaluate the model's generalisation capabilities. For three specific configurations, we created entirely new non-PRODIGy characters, not found in any existing resource, to assess the model's ability to generalise when dealing with completely novel and unseen personas.

The configurations are designed to vary across two key dimensions—\textit{interlocutors} and \textit{topics}—as outlined in~Q2:

\begin{itemize}[leftmargin=*]
    \item \textbf{Speaker pairings}: Dialogues were generated using the following configurations:
    \begin{enumerate}[label=(\roman*), leftmargin=*]
        \item \textit{Familiar pairings}: Characters were paired (a) as they appear in the PRODIGy dataset (e.g., Harry and Sally from the movie \textit{When Harry Met Sally}), (b) using entirely new, non-PRODIGy characters (e.g., John Doe and Jane Doe).\footnote{We treat new characters as familiar because the unfamiliar category comprises cases where we forced a highly unexpected interlocutor onto a speaker, which is not the case for new characters as there are no expectations.}
        \item \textit{Unfamiliar pairings}: Characters from different movies within the PRODIGy dataset were paired (e.g., Harry and Superman), or PRODIGy characters were matched with new non-PRODIGy characters (e.g, Harry and John Doe).
    \end{enumerate}

    \item \textbf{Topics}: Dialogues were generated using the following types of topics:
    \begin{enumerate}[label=(\roman*), leftmargin=*]
        \item \textit{Familiar topics}: Those associated with the characters in the PRODIGy dataset (e.g., Harry and Sally discussing about love).
        \item \textit{Unfamiliar topics}: New topics unfamiliar to the target speaker (e.g., Harry having a dialogue about war).
    \end{enumerate}
\end{itemize}

\noindent Further details are provided in Appendix~\ref{experiments-details}.

\paragraph{Non-PRODIGy Characters Generation} To generate the new non-PRODIGy characters, we employed the GPT4o-mini \cite{openai2024gpt4omini} model and the Persona Hub dataset \cite{ge2024scaling}, comprising 1 billion personas automatically extracted and curated from web data.
We selected GPT4o-mini based on a preliminary experiment where it demonstrated the ability to generate detailed profiles while remaining cost-efficient. From the Persona Hub dataset, we randomly selected 100 personas and crafted detailed prompts using these persona descriptions (see Appendix~\ref{profile-generation}). These prompts were used to instruct GPT4o-mini to generate biographies in the style of PRODIGy biographies. Unlike the original PRODIGy biographies, consisting of fictional movie characters, the newly generated biographies represent ordinary persons.

\paragraph{Dialogue Generation} 

We approached dialogue generation as a next-turn generation task.
The prompt was updated at each step to include the dialogue history up to that point. 
To maintain consistency, we limited each character's biography to their first five sentences across all configurations, as this represented the minimum biography sentence number in PRODIGy. Each dialogue was generated with a total of 8 turns, in line with recent findings showing that persona alignment tends to decay beyond this point, with noticeable “instruction drift” occurring after 8 turns \cite{li2024measuringcontrollinginstructioninstability}. This choice ensures we capture the phase where persona signals are strongest and facilitates more consistent human evaluation. We generated the dialogues starting from each combination of speakers, thus obtaining a total of 4375 dialogues. Figure~\ref{fig:dialogue_esempio} shows an example of generated dialogue. 

\subsection{Evaluation} 

A model can be considered successful in generating responses that align with a target biography and adapt to the interlocutor's information, if the speakers' biographies are identifiable from the generated dialogue. Recognising profiling indicators in text has been shown to be a complex task, especially for humans \cite{youyou2015computer, flekova-etal-2016-analyzing, de-mattei-etal-2020-invisible}. To address this challenge, we employed an LLM-as-a-judge approach in addition to human evaluations. Moreover, we framed the task as author identification, where given a dialogue and three possible biographies, the evaluators (both human and automatic) were tasked to identify which biography belongs to the target speaker. While traditional statistical classifiers could serve as an additional baseline, we chose not to include them, as such models typically rely on surface-level textual features, which are not well-suited to assess the models’ ability to adapt to speakers’ personas.  In contrast, our framework, grounded in both human judgment and LLM evaluation, is designed to assess deeper aspects of persona consistency that a statistical method might overlook.

\begin{figure*}[h]
    \centering
    \includegraphics[width=1\textwidth]{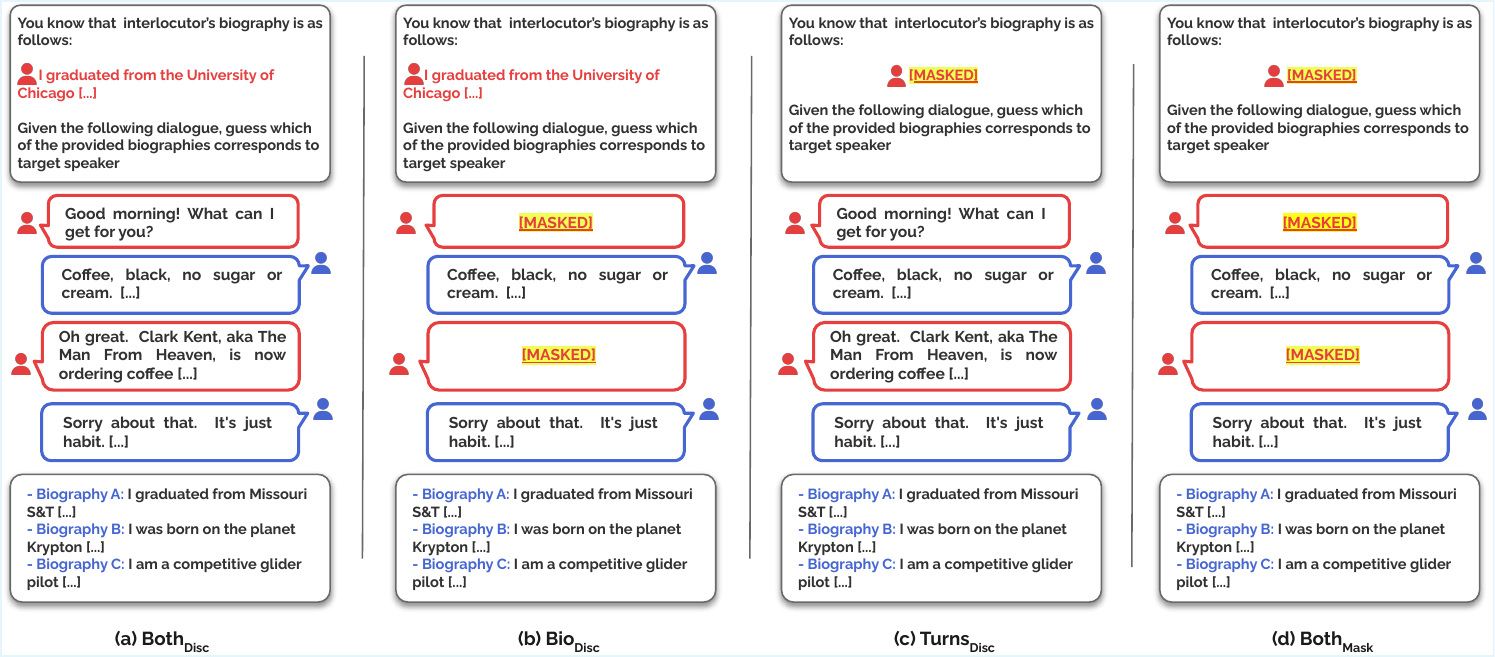}
  \caption{
Information disclosure configurations used in the evaluation: (a) Both$_{Disc}$: both the interlocutor’s biography and turns are visible to the evaluator; (b) Bio$_{Disc}$: only the interlocutor’s biography is visible, while their turns are masked in the dialogue; (c) Turns$_{Disc}$: only the interlocutor’s turns are visible, with their biography masked; (d) Both$_{Mask}$: both the interlocutor’s biography and turns are masked. The illustrated examples are simplified for clarity and demonstrate the input available to evaluators in each condition.}
    \label{fig:disclosure-configs}
\end{figure*}

We designed four distinct configurations, varying in the degree of interlocutor information disclosed to the evaluator (see Figure~\ref{fig:disclosure-configs} for examples): 

\begin{enumerate}[leftmargin=*]
    \item \textbf{Interlocutor's biography and turns disclosed (Both$_{Disc}$)}: Evaluators were provided with the full dialogue, the interlocutor's biography and dialogue turns. This configuration examines whether the combination of the interlocutor's information with the target speaker's turns enhances the ability to accurately recognise the correct target speaker's biography (Figure~\ref{fig:disclosure-configs}a).

    \item \textbf{Interlocutor's biography disclosed and turns masked (Bio$_{Disc}$)}: Evaluators received the interlocutor's biography, but their dialogue turns were hidden. This setup tests whether the interlocutor's biography is more relevant than their turns in order to correctly recognise the target speaker's biography (Figure~\ref{fig:disclosure-configs}b).

    \item \textbf{Interlocutor's biography masked and turns disclosed (Turns$_{Disc}$)}: Evaluators were given the full dialogue with the interlocutor's biography hidden. This configuration examines whether the interlocutor's dialogue turns alone are more relevant than their biography for accurately identifying the target speaker's biography (Figure~\ref{fig:disclosure-configs}c).

    \item \textbf{Both interlocutor's biography and turns masked (Both$_{Mask}$)}: Neither the interlocutor's biography nor their dialogue turns were provided. By masking all interlocutor information, we assess whether the target speaker's biography can still be accurately identified without any clue about the other speaker (Figure~\ref{fig:disclosure-configs}d).

\end{enumerate}

\noindent To make recognition challenging and prevent evaluators from relying on overly obvious cues in the biographies, we selected the two biographies most semantically similar to the target speaker's biography as alternative options. For this, we used  SBERT \cite{thakur-2020-AugSBERT}, computing the cosine similarity of the biography embeddings generated with the all-MiniLM-L6-v2 model\footnote{\url{https://huggingface.co/sentence-transformers/all-MiniLM-L6-v2}}. Additionally, we crafted four instruction templates to handle the four disclosure configurations (see~Appendix~\ref{judge-llm-prompts}).

\paragraph{Automatic Evaluation}

We fine-tuned Llama 3.1 8B Instruct as LLM-as-a-judge utilising the dialogue partitions from the PRODIGy dataset as described in Section~\ref{dataset}. We opted for a fine-tuned model to specialise the evaluator for the domain of movie dialogues, which exhibit distinctive stylistic and structural features that our dialogue model consistently manages to emulate. At inference time, we employed a greedy decoding mechanism to generate the LLM-as-a-judge's predictions, ensuring consistency and determinism in the evaluation process \cite{song2024good}. Our methodology involved automatic assessment of both gold and generated dialogues, with each dialogue evaluated twice, once for the target speaker and once for the interlocutor. 

\paragraph{Human Evaluation}

For high-quality human evaluation, we recruited native English speakers from Prolific\footnote{\url{https://www.prolific.com/}}, a platform specifically designed for research experiments, granting fair payment to annotators. 
Each dialogue was evaluated by three evaluators, who conducted a detailed review of the dialogues across the four disclosure configurations, following the same assessment methodology as the LLM-as-a-judge to maintain consistency in evaluation criteria. Additionally, we held a post-hoc qualitative interview with the evaluators to gain further insights into the evaluation. The evaluation guidelines are reported in Appendix \ref{human-eval-appendix}.

\section{Automatic Results}

In this section, we present the LLM-as-a-judge's performance on the gold test set and on model-generated dialogues, including both the fine-tuning and zero-shot configurations. Results are shown for the various biography- and turn-disclosure conditions. We employ the following metrics: accuracy (Acc), precision (Prec), recall (Rec), and F1 Score (F1). Detailed results are reported in Appendix \ref{detailed-automatic-results}.

\paragraph{Baseline}
As a baseline, we take a random guess over the three speakers, yielding an accuracy of 0.333 across all configurations, as the pool of possible candidates for speaker identification is always three under all settings.

\subsection{Gold Dialogues}

Table~\ref{tab:gold-classifier-disclosure} examines the effect of various degrees of disclosure of the interlocutor's information on the judge's ability to correctly identify speakers. We observe that providing the LLM-as-a-judge with the interlocutor's biographical information significantly enhances the judge's accuracy. This suggests that in the gold data the interlocutor's information is relevant to the judge LLM. When both the interlocutor’s biography and their turns are provided (Both$_{Disc}$), the model achieves a high accuracy of 0.820. Providing the biography only (Bio$_{Disc}$) still supports robust performance, with a comparable accuracy of 0.805.
Masking the biography instead yields a notable drop: relying solely on the interlocutor’s disclosed turns (Turns$_{Disc}$) lowers the performance (0.588 accuracy), while masking both the biography and turns (Both$_{Mask}$) leads to further reductions (0.577 accuracy).

\begin{table}[ht!]
\small
\centering
\begin{tabular}{lrrrr}
\toprule
\textbf{Disclosure} & \textbf{Acc} & \textbf{Prec} & \textbf{Rec} & \textbf{F1} \\ \midrule
Both$_{Disc}$ & \textbf{0.820} & \textbf{0.838} & \textbf{0.817} & \textbf{0.819} \\
Bio$_{Disc}$  & 0.805          & 0.822          & 0.801          & 0.802          \\
Turns$_{Disc}$ & 0.588         & 0.629          & 0.583          & 0.581          \\
Both$_{Mask}$  & 0.577         & 0.619          & 0.571          & 0.568          \\ \bottomrule
\end{tabular}
\caption{LLM-as-a-judge results on PRODIGy gold dialogues across various degrees of interlocutor information disclosure.}
\label{tab:gold-classifier-disclosure}
\end{table}

\noindent The performance of the LLM-as-a-judge on gold dialogues clearly demonstrates its ability to accurately identify speakers. Moreover, the analysis highlights that biographical information plays a critical role in significantly enhancing the model's speaker recognition capabilities.

\subsection{Fine-Tuned Model's Generations}

In this section, we present the results of the evaluation of the fine-tuned model's dialogues. 

\paragraph{The fine-tuned model effectively adapts its responses to both the target and interlocutor speakers ~(Q1)}
As shown in Table~\ref{tab:disclosure_metrics}, the interlocutor's biography plays a more significant role than their conversational turns. This indicates that the model efficiently tailors its responses to the interlocutor's biography. Notably, the performances for conditions where both the interlocutor's biography and turns are disclosed (\textit{Both$_{Disc}$}) and where only the biography is disclosed (\textit{Bio$_{Disc}$}) are comparable (0.594 accuracy). 
Instead, performance drops significantly when only interlocutor's turns are disclosed (\textit{Turns\textsubscript{Disc}}) or both interlocutor's  turns and biography are masked (\textit{Both\textsubscript{Mask}}) (accuracy of 0.517 and 0.515, respectively). These lower scores, compared to gold dialogues, could also be due to a data contamination effect, as the judge LLM might be familiar with the gold dialogues.

\begin{table}[ht]
    \centering
    \small
    \begin{tabular}{lrrrr}
        \toprule
        \textbf{Disclosure}       & \textbf{Acc} & \textbf{Prec} & \textbf{Rec} & \textbf{F1} \\
        \midrule
        Both$_{Disc}$            & \textbf{0.594}        & \textbf{0.618}         & 0.594        & 0.588       \\
        Bio$_{Disc}$           & \textbf{0.594}        & 0.615         & \textbf{0.595}        & \textbf{0.589}       \\
        Turns$_{Disc}$         & 0.517        & 0.552         & 0.518        & 0.508       \\
        Both$_{Mask}$                & 0.515        & 0.546         & 0.515        & 0.505       \\
        \bottomrule
    \end{tabular}
    \caption{LLM-as-a-judge results on fine-tuned model's dialogues across different degrees of interlocutor information disclosure.}
    \label{tab:disclosure_metrics}
\end{table}

These results highlight the importance of the interlocutor's information in enabling evaluators to correctly match the target speaker's biography, suggesting that the model appropriately adapts its responses to both the target and interlocutor.

\paragraph{The fine-tuned model adapts well to familiar speaker pairings but struggles with unfamiliar ones ~(Q2)}

Table~\ref{tab:speaker_metrics} highlights that the LLM-as-a-judge performs best in dialogues with familiar speaker pairings, such as PRODIGy-PRODIGy pairs from the same movie or non-PRODIGy-non-PRODIGy pairs (0.703 accuracy). 
This improved performance likely stems from the pairings aligning with the LLM's prior knowledge, leading to more intuitive dialogues.
Conversely, the model's performance drops substantially with unfamiliar speaker pairings, such as PRODIGy-PRODIGy pairs from different movies or PRODIGy-non-PRODIGy pairings (0.496 accuracy), indicating that these dialogues are more challenging for speaker identification.

\begin{table}[ht]
    \centering
    \small
    \begin{tabular}{lrrrr}
        \toprule
        \textbf{Speakers}         & \textbf{Acc}   & \textbf{Prec}  & \textbf{Rec}   & \textbf{F1}    \\
        \midrule
        Familiar      & \textbf{0.703} & \textbf{0.728} & \textbf{0.701} & \textbf{0.699}\\
        Unfamiliar  & 0.496 & 0.524 & 0.498 & 0.487 \\
        \bottomrule
    \end{tabular}
    \caption{LLM-as-a-judge results on fine-tuned model's generated dialogues across different speaker pairings.}
    \label{tab:speaker_metrics}
\end{table}

\noindent Overall, dialogues generated from familiar speaker pairs make it easier to identify the target speaker's biography, while unfamiliar pairings prove less intuitive, leading to a lower generalisation ability.

\paragraph{The fine-tuned model effectively generalises across topics (Q2)}

Table~\ref{tab:topic_metrics} shows that the LLM maintains relatively stable performance across different topics, with performance only slightly higher for the familiar topic (0.566 vs 0.561 accuracy).

\begin{table}[ht]
    \centering
    \small
    \begin{tabular}{lrrrr}
        \toprule
        \textbf{Topic} & \textbf{Acc} & \textbf{Prec} & \textbf{Rec} & \textbf{F1} \\
        \midrule
        Familiar       & \textbf{0.566}       & \textbf{0.595}         & \textbf{0.566}       & \textbf{0.558}      \\
        Unfamiliar         & 0.561        & 0.588         & 0.563        & 0.554       \\
        \bottomrule
    \end{tabular}
    \caption{LLM-as-a-judge results on fine-tuned model's dialogues across topics.}
    \label{tab:topic_metrics}
\end{table}

\noindent The results reveal no significant differences in performance between dialogues with familiar topics and those with unfamiliar topics, suggesting robust generalisation across topics instead.

\subsection{Zero-Shot Generations}

The results in this section are based on dialogues generated in a zero-shot setting. 
The findings are consistent with those observed for the gold dialogues and the fine-tuned model's generations. 

\paragraph{In a zero-shot setting, the model effectively tailors its responses to both  target speaker and interlocutor ~(Q1)}
Table~\ref{tab:zero_shot_disclosure} shows results for various interlocutor's information disclosure settings. As for the fine-tuned model, performance is highest when both biography and turns are disclosed (\textit{Both\textsubscript{Disc}}, 0.755 accuracy). Partial disclosure -- either only biography (\textit{Bio$_{Disc}$}) or turns (\textit{Turns$_{Disc}$}) -- yields slightly lower metrics, while masking both (\textit{Both$_{Mask}$}) yields the lowest accuracy.

\begin{table}[ht]
    \centering
    \small
    \begin{tabular}{lrrrr}
        \toprule
        \textbf{Disclosure}   & \textbf{Acc} & \textbf{Prec} & \textbf{Rec} & \textbf{F1} \\
        \midrule
        Both$_{Disc}$        & \textbf{0.755}        & \textbf{0.769}         & \textbf{0.754}        & \textbf{0.754}       \\
        Bio$_{Disc}$        & 0.751        & 0.763         & 0.749        & 0.749       \\
        Turns $_{Disc}$           & 0.742        & 0.771         & 0.740        & 0.740       \\
        Both$_{Mask}$              & 0.735        & 0.757         & 0.734        & 0.733       \\
        \bottomrule
    \end{tabular}
    \caption{LLM-As-A-Judge performance metrics for different disclosure types in zero-shot dialogues.}
    \label{tab:zero_shot_disclosure}
\end{table}

\noindent  These findings underscore that the dialogue model adapts appropriately to the interlocutor's information, allowing for better speaker recognition.

\paragraph{In a zero-shot setting, the model adapts well to familiar speaker pairings but struggles with unfamiliar ones ~(Q2)}

Table~\ref{tab:zero_shot_dialogues} shows that dialogues involving familiar speaker pairs yield better performance metrics (e.g., 0.816 accuracy).
In contrast, dialogues between unfamiliar speaker pairings lead to a marked drop (0.696 accuracy). 
This pattern mirrors the trends observed in the fine-tuned model, emphasising the importance of familiar speaker pairings for coherent dialogue generation.
\begin{table}[ht]
    \centering
    \small
    \begin{tabular}{lrrrr}
        \toprule
        \textbf{Speakers}         & \textbf{Acc} & \textbf{Prec} & \textbf{Rec} & \textbf{F1} \\
        \midrule
        Familiar      & \textbf{0.816} & \textbf{0.831} & \textbf{0.816} & \textbf{0.815} \\
        Unfamiliar  & 0.696 & 0.718 & 0.695 & 0.694 \\
        \bottomrule
    \end{tabular}
    \caption{LLM-as-a-judge results on zero-shot
generated dialogues across different speaker pairings.}
    \label{tab:zero_shot_dialogues}
\end{table}

\paragraph{In a zero-shot setting, the model effectively generalises across topics ~(Q2)}
Table~\ref{tab:zero_shot_topic} shows a consistent performance of the LLM across familiar and unfamiliar topics. This trend, consistent with that of the fine-tuned model, suggests that the LLM can robustly generalise in a zero-shot setting.

\begin{table}[ht]
    \centering
    \small
    \begin{tabular}{lcccc}
        \toprule
        \textbf{Topic} & \textbf{Acc} & \textbf{Prec} & \textbf{Rec} & \textbf{F1} \\
        \midrule
        Familiar       & \textbf{0.737}        & 0.755         & \textbf{0.736}        & \textbf{0.735}       \\
        Unfamiliar         & 0.735        & \textbf{0.756}         & 0.735        & 0.734       \\
        \bottomrule
    \end{tabular}
    \caption{LLM-As-A-Judge performance for zero-shot dialogues across topics.}
    \label{tab:zero_shot_topic}
\end{table}

\subsection{Fine-Tuning vs Zero-Shot}

Notably, the performance metrics for zero-shot dialogues are higher than for the fine-tuned model. We hypothesise that this may be due to the LLMs tendency to replicate information directly from their prompts \cite{russo-etal-2023-countering, casula-etal-2024-delving}.  In our scenario, the model might replicate in the generations details from the provided biographies. While facilitating speaker recognition, this "copy-paste" tendency may also limit the model's ability to fully capture and convey deeper characteristics of speakers, resulting in outputs that merely copy the surface elements of the biographical data. To investigate this hypothesis, we conducted an additional analysis examining word overlaps 
between the speakers' biographies and their turns.

\paragraph{Additional fine-tuning on persona-based dialogues is beneficial to the model ~(Q3)}

The results revealed a pronounced tendency for the zero-shot setting to reuse specific terms directly from the biographies. Checking rare words from the biographies, we found that they appeared up to eight times more frequently in zero-shot dialogues (40.69\%) compared to those generated by the fine-tuned model (5.18\%) and to the gold dialogues (5.84\%). Details are in Table~\ref{tab:rare-word-overlap} in Appendix~\ref{overlap-metrics}.

\noindent This "copy-paste" behaviour makes the speaker more easily recognisable, but also the outputs more superficial. In contrast, the fine-tuned model's generations seem to resemble more the gold dialogues, indicating a stronger ability to generate responses reflecting deeper speaker characteristics.\footnote{A similar phenomenon was discussed in \cite{occhipinti-etal-2024-prodigy}, where models fine-tuned on data with high overlap between dialogues and biographies resulted in trivial "copy-pasting" models.}

\begin{figure}[!t]
\centering
\includegraphics[width=1\columnwidth]{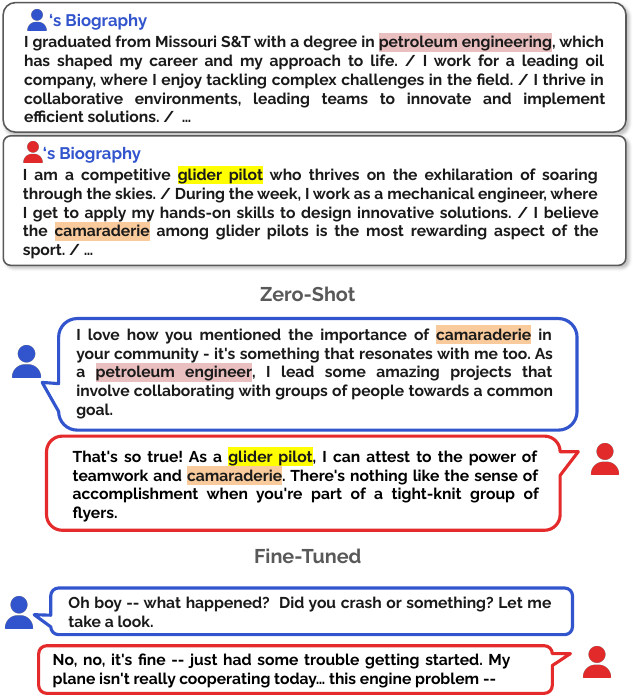}
 \caption{Comparison of dialogues generated in a zero-shot setting vs. the fine-tuned model, using the same input and Non-PRODIGy characters.}
 \label{fig:dialogue_overlap}
 \end{figure}

Figure~\ref{fig:dialogue_overlap} illustrates this phenomenon. The biographies describe two distinct personas: one is a \textit{petroleum engineer} who thrives in collaborative problem-solving environments, while the other is a \textit{glider pilot} who values \textit{camaraderie}. While the dialogue from the zero-shot setting merely reproduces these exact terms instead of more common alternatives (e.g., \textit{friendship among pilots}), the fine-tuned models' dialogue reflects deeper persona traits by portraying a collaborative and problem-solver person speaking to a pilot facing engine issues.

Notably, while looking at infrequent words clearly shows the ``copy-paste" phenomenon, this is not limited to rare word repetition. We also observe significantly higher BLEU, ROUGE, and METEOR scores when comparing biographies and dialogue turns in the zero-shot setting compared to the fine-tuned model (e.g., the average METEOR score for the fine‑tuned model is 0.043, whereas it is 0.140 for the zero‑shot setting; see Table~\ref{tab:overlap-metrics} in Appendix~\ref{overlap-metrics}). This indicates that copying extends to broader content overlap, not just rare words. 

\section{Human Evaluation Results}

The human evaluation, conducted by evaluators hired through Prolific, consisted of 360 assessments focusing on a subset of dialogues previously evaluated by the LLM-as-a-judge. Dialogues were randomly selected and stratified by experiment type and level of interlocutor information disclosure to ensure balanced representation. Due to the "copy-paste" phenomenon observed in zero-shot dialogues, which could make the task trivial, the evaluation was limited to gold dialogues and dialogues generated by the fine-tuned model.

Overall, the trends observed in the human evaluation align with those from the LLM-as-a-judge assessment. 
However, scores from the human evaluation are considerably lower than those for the LLM-as-a-judge, indicating that LLMs surpass human capabilities in this task. This observation is consistent with previous findings 
\cite{youyou2015computer, flekova-etal-2016-analyzing, de-mattei-etal-2020-invisible}. 

\paragraph{Human evaluations confirm that the model effectively adapts its responses to target and interlocutor speakers (Q1)}
Human evaluation results show that providing the interlocutor's biography significantly improves biography recognition for both generated and gold dialogues. In contrast, masking both the interlocutor's biography and turns severely hinders the evaluators' ability to accurately identify the correct biography.

Table~\ref{tab:human_eval_disclosure} presents the human evaluation results across the four levels of interlocutor information disclosure. Disclosing both interlocutor's biography and turns leads to the most accurate identification of the target speaker, as indicated by the highest accuracy (0.509 in Both$_{Disc}$). Notably, revealing only the interlocutor’s turns while masking their biography results in better outcomes (0.474 accuracy in Turns$_{Disc}$) compared to providing the interlocutor's biography while masking their turns (0.449 accuracy in Bio$_{Disc}$). The lowest accuracy is observed when all interlocutor information is masked (0.351 accuracy in Both$_{Mask}$).

\begin{table}[ht]
    \centering
    \small
    \begin{tabular}{lrrrr}
        \toprule
        \textbf{Disclosure}       & \textbf{Acc} & \textbf{Prec} & \textbf{Rec} & \textbf{F1} \\
        \midrule
        Both$_{Disc}$            & \textbf{0.509}        & \textbf{0.478}         & 0.375        & \textbf{0.395}       \\
        Bio$_{Disc}$             & 0.449        & 0.447         & \textbf{0.388}        & 0.370       \\
        Turns$_{Disc}$           & 0.474        & 0.434         & 0.385        & 0.375       \\
        Both$_{Mask}$            & 0.351        & 0.332         & 0.248        & 0.249       \\
        \bottomrule
    \end{tabular}
    \caption{Human evaluation results on fine-tuned model's generations and gold dialogues across different degrees of interlocutor information disclosure.}
    \label{tab:human_eval_disclosure}
\end{table}

\paragraph{Human evaluations confirm that the model better adapts its responses with familiar speaker pairings, while struggles with unfamiliar ones (Q2)}
Evaluators achieve the highest performance when dealing with dialogues involving common speaker pairings. 
The results in Table~\ref{tab:human_eval_common} indicate that generating dialogues from common speaker pairings leads to higher target speaker recognition (0.497 accuracy), 
whereas unfamiliar pairings result in lower performance (0.373 accuracy).

\begin{table}[ht]
    \centering
    \small
    \begin{tabular}{lrrrr}
        \toprule
        \textbf{Speakers}         & \textbf{Acc} & \textbf{Prec} & \textbf{Rec} & \textbf{F1} \\
        \midrule
        Familiar      & \textbf{0.497} & \textbf{0.512} & \textbf{0.440} & \textbf{0.419} \\
        Unfamiliar  & 0.373 &	0.336 &	0.264 &	0.273 \\
        \bottomrule
    \end{tabular}
    \caption{Human evaluation results on fine-tuned
generated dialogues across different speaker pairings.}
    \label{tab:human_eval_common}
\end{table}

A post-hoc interview with the evaluators revealed that they were unable to differentiate between the generations and the gold dialogues, suggesting that the model’s outputs closely mirrored the style and characteristics of the gold dialogues. The evaluators also reported that identifying the correct biography was challenging in most cases, as multiple biographies could plausibly match the target's turns. In many instances, the choice was determined only by spotting a small detail.

\section{Conclusion}

We studied the role of interlocutor information in persona-based dialogue generation, analysing how LLMs adapt their responses based on both target speaker and interlocutor characteristics. We introduced a novel, systematic framework for evaluating the model’s ability to (i) align responses with both target and interlocutor personas; (ii) its generalisation abilities across familiar and unfamiliar speaker pairings and across topics familiar and unfamiliar to the target speaker; (iii) and the impact of fine-tuning compared to zero-shot generation. 
Our findings reveal that while models effectively align responses with target speakers and adapt to interlocutor biographies, their performance drops with unfamiliar pairings. Similarly, models demonstrate strong generalisation across different topics, even those unfamiliar to the target speaker. Notably, dialogues generated in a zero-shot setting achieve higher speaker recognition accuracy by directly "copy-pasting" biographical details, thereby also leading to much more superficial dialogues. In contrast, fine-tuned models more accurately capture deeper speaker traits. 
These insights have practical relevance for personalised dialogue systems, such as digital assistants, chatbots, and social robots, where maintaining a consistent and contextually appropriate persona is essential for enhancing user trust, engagement, and interaction.

\section*{Limitations}

Our experiments rely on the PRODIGy dataset, which features movie dialogues and fictional characters. This choice may introduce stereotyped roles and inherent biases. Additionally, while our evaluation combines LLM-as-a-judge and human assessments, each method has limitations: the LLM evaluator might be affected by data contamination, and human evaluations are subject to individual interpretation and limited reproducibility.

\section*{Ethics Statements}

Role-play agents carry several risks, including toxicity, bias, hallucinations, and privacy violations \cite{chen2024persona}. Early studies have revealed a tendency of these models to generate harmful content \cite{wen-etal-2023-unveiling}, which not only degrades the user experience but also presents significant safety concerns. Furthermore, they often exhibit role-based biases, stemming from both inherent biases in their pre-training data \cite{xue2023bias} and user prompts that may inadvertently direct them toward biased outputs \cite{perez2022ignore, branch2022evaluating}. Additionally, these systems might exhibit \textit{character hallucination} \cite{ahn-etal-2024-timechara}, i.e., producing responses that do not align with their assigned roles. Lastly, role-play systems may pose privacy risks by inadvertently disclosing users' private information \cite{krishnamurthy2011privacy, corrigan2014does}.

\bibliography{custom}

\appendix

\section{Dialogue Annotation With Topic}\label{topic-annotation}
In this section, we detail the dialogue annotation process for topic labelling. 

First, each dialogue was assigned three candidate topics using Llama 3.1 8B Instruct \cite{dubey2024llama}\footnote{We excluded 10 dialogues with harmful content, as the model declined to assign topics to them.}. Next, we extracted the most frequent topics by analysing the word stems in these labels and selected the 100 most common stems. These stems were then clustered into broader thematic categories, and each dialogue was assigned the corresponding cluster label based on its generated topics. Finally, we validated the labelling accuracy by manually reviewing and correcting a sample of 200 dialogues. The complete procedure is outlined in Algorithm~\ref{alg:topic_annotation}.

\begin{algorithm}
\small
\begin{algorithmic}[1]
    \Function{AnnotateTopics}{$Dialogues$}
        \ForAll{$d \in Dialogues$}
            \State $T_d \gets$ \Call{GenerateTopics}{$d$, 3} 
        \EndFor
        \State $W \gets$ \Call{ExtractStems}{$\bigcup_{d \in Dialogues} T_d$}
        \State $S \gets$ \Call{SelectTop}{$W$, 100}
        \State $\mathcal{C} \gets$ \Call{ClusterStems}{$S$}
        \ForAll{$d \in Dialogues$}
            \State $d.label \gets$ \Call{AssignClusterLabel}{$T_d, \mathcal{C}$}
        \EndFor
        \State \Call{HumanValidationLabels}{$Dialogues$, 200}
    \EndFunction
\end{algorithmic}
\caption{Topic Annotation}
\label{alg:topic_annotation}
\end{algorithm}

\section{Model Fine-Tuning and Generations}\label{model-generation-appendix}
We fine-tuned the Llama 3.1 8B Instruct using QLoRA \cite{dettmers2023qlora}, with learning rate 3e-4, low-rank approximation set to 16, low-rank adaptation
set to 8, and dropout rate set to 0.0\footnote{Followed hyper-parameters as \url{https://github.com/pytorch/torchtune/blob/main/recipes/configs/llama3_1/8B_qlora_single_device.yaml}}. Evaluation
steps were set at 524, batch size at 4 and gradient
accumulation step at 2. We employed 10 epochs
with early stopping after 3 epochs.

For generation, we set top-p to 0.9, temperature to 0.8, and applied a repetition penalty of 1.2.

We framed both processes as a next-turn generation task. Dialogue data was introduced incrementally, turn by turn, allowing the model to better capture and adapt to conversational context. 

\section{Prompts}
\label{sec:appendix}

\subsection{Non-PRODIGy Character Generation Prompt}\label{profile-generation}

To generate biographies in the style of PRODIGy, we used the prompt shown in Table~\ref{tab:prompt-new-characters} to guide GPT4o-mini. Each persona from the Persona Hub dataset was unfamiliarly assigned a gender and a Myers-Briggs Type Indicator (MBTI) personality type. To ensure alignment with PRODIGy biographies, we instructed GPT4o-mini to include details about the individual's job, relationship status, lifestyle, and family background.

\prompttable{1}{Considering the following persona sentence: \\\\

        \textbf{\texttt{[persona sentence]}}\\\\

        the following gender: \\\\

        \textbf{\texttt{[gender]}}\\\\

        and the following mbti: \\\\

        \textbf{\texttt{[mbti]}}\\\\

        please create a profile of a person, using the following structure:\\\\

        \textbf{\texttt{\string{\string{"gender": "gender",
        "mbti": "mbti",
        "biography": [
        "sentence 1",
         "sentence 2", 
         "...",
        "sentence 10"
        ]\string}\string}}}\\\\

        Please ensure the biography contains up to 10 sentences in the first person singular. Include details about the individual’s job, relationship status, lifestyle, and family background. Be sure to capture a varied portrayal of the individual's life and character.

        Please ensure to provide the dictionary only, without anything else.}{Prompt to generate Non-PRODIGy biographies}{prompt-new-characters}

\subsection{Generation Prompt}

Table~\ref{tab:generation-prompt} shows the prompt employed for generations, both in fine-tuning and zero-shot settings.

\prompttable{2}{You are \textbf{\texttt{[target speaker]}}. Your biography is as follows:\\\\

\texttt{\textbf{[biography sentences]}}\\\\

You are having a dialogue with \texttt{\textbf{[interlocutor speaker]}}. Their biography is as follows: \\\\

\texttt{\textbf{[biography sentences]}}\\\\

You are discussing about the topic \texttt{\textbf{[topic]}}. Considering the dialogue history, provide an answer to \texttt{\textbf{[interlocutor speaker]}}. Be sure to provide the answer only.\\\\ 

\texttt{\textbf{[dialogue history]}}}{Prompt for both fine-tuning and dialogue generation}{generation-prompt}

\subsection{Judge LLM prompts}\label{judge-llm-prompts}

In this section, we showcase the four prompts employed for the LLM-as-a-Judge evaluation. Table~\ref{tab:Both-disc-eval-prompt} displays the prompt for the Both$_{Disc}$ configuration, where both the interlocutor's biography and turns are disclosed. Table~\ref{tab:Bio-Disc-eval-prompt} corresponds to the Bio$_{Disc}$ setting, where only the interlocutor’s biography is disclosed while their turns are masked. Similarly, the Turns$_{Disc}$ configuration (Table~\ref{tab:Turns-Disc-eval-prompt}) masks the interlocutor’s biography while disclosing their turns. Finally, Table~\ref{tab:Both-Mask-eval-prompt} represents the Both$_{Mask}$ condition, where both the interlocutor’s biography and turns are masked.

\prompttable{3}{You know that \textbf{\texttt{interlocutor}}'s biography is as follows: \\\\

        \textbf{\texttt{[interlocutor's biography]}}\\\\

        Given \textbf{\texttt{interlocutor}}'s biography and the following dialogue about the topic \textbf{\texttt{[topic]}}, your task is to guess which of the provided biographies corresponds to \textbf{\texttt{target speaker}}. 
        The biographies are provided at the end of the dialogue. 
        Please provide your answer as "Biography A", "Biography B", or "Biography C". 
        Please make your guess even though the dialogue may sound a little weird or unnatural. 
        Your response must follow this JSON format:
        \texttt{\{"Guess": "Biography X"\}} \\\\

        DIALOGUE \\\\

        \textbf{\texttt{[Dialogue]}} \\\\

        BIOGRAPHIES \\\\

        Biography A:
        \textbf{\texttt{[Biography A sentences]}}

        Biography B:
        \textbf{\texttt{[Biography B sentences]}}

        Biography C:
        \textbf{\texttt{[Biography C sentences]}}}{Both interlocutor's biography and turns disclosed}{Both-disc-eval-prompt}

\prompttable{4}{You know that \textbf{\texttt{interlocutor}}'s biography is as follows: \\\\

        \textbf{\texttt{[interlocutor's biography]}} \\\\

        Given \textbf{\texttt{interlocutor}}'s biography and the following dialogue about the topic \textbf{\texttt{[topic]}}, in which \texttt{interlocutor}'s turns are masked, your task is to guess which of the provided biographies corresponds to \textbf{\texttt{target speaker}}. 
        The biographies are provided at the end of the dialogue. 
        Please provide your answer as "Biography A", "Biography B", or "Biography C". 
        Please make your guess even though the dialogue may sound a little weird or unnatural. 
        Your response must follow this JSON format:
        \texttt{\{"Guess": "Biography X"\}}\\\\

        DIALOGUE\\\\

        \textbf{\texttt{[Dialogue]}}\\\\

        BIOGRAPHIES\\\\

        Biography A:
        \textbf{\texttt{[Biography A sentences]}}

        Biography B:
        \textbf{\texttt{[Biography B sentences]}}

        Biography C:
        \textbf{\texttt{[Biography C sentences]}}}{Interlocutor's biography disclosed, turns masked}{Bio-Disc-eval-prompt}

\prompttable{5}{Given the following dialogue about the topic \texttt{topic}, your task is to guess which of the provided biographies corresponds to \texttt{target speaker}. 
        The biographies are provided at the end of the dialogue. 
        Please provide your answer as "Biography A", "Biography B", or "Biography C". 
        Please make your guess even though the dialogue may sound a little weird or unnatural. 
        Your response must follow this JSON format:
        \texttt{\{"Guess": "Biography X"\}}\\\\

        DIALOGUE \\\\

        \textbf{\texttt{[Dialogue]}}\\\\

        BIOGRAPHIES\\\\

        Biography A:
        \textbf{\texttt{[Biography A sentences]}}

        Biography B:
        \textbf{\texttt{[Biography B sentences]}}

        Biography C:
        \textbf{\texttt{[Biography C sentences]}}}{Interlocutor's biography masked, turns disclosed}{Turns-Disc-eval-prompt}

\prompttable{6}{Given the following dialogue about the topic \texttt{topic}, in which \texttt{interlocutor}'s turns are masked, your task is to guess which of the provided biographies corresponds to \texttt{target speaker}. 
        The biographies are provided at the end of the dialogue. 
        Please provide your answer as "Biography A", "Biography B", or "Biography C". 
        Please make your guess even though the dialogue may sound a little weird or unnatural. 
        Your response must follow this JSON format:
        \texttt{\{"Guess": "Biography X"\}}

        DIALOGUE\\\\

        \textbf{\texttt{[Dialogue]}}\\\\

        BIOGRAPHIES\\\\

        Biography A:
        \textbf{\texttt{[Biography A sentences]}}

        Biography B:
        \textbf{\texttt{[Biography B sentences]}}

        Biography C:
        \textbf{\texttt{[Biography C sentences]}}}{Both interlocutor's biography and turns masked}{Both-Mask-eval-prompt}

\section{Experimental Design Details}\label{experiments-details}

This section outlines our experimental settings, as detailed in Table~\ref{tab:experiments_overview}. We designed seven experiments to assess how speaker pairing and topic variation affect dialogue generation, introducing incremental randomness to control for the data contamination effect. We label a configuration as \textit{familiar} when the speakers belong to the same movie or belong to the same context (i.e., Non-PRODIGy characters), and as \textit{unfamiliar} when the speakers belong to atypical contexts (e.g., characters coming from different movies or movie characters talking to ordinary persons).
In Exp1, we preserve both the original PRODIGy pairing (target P\textsubscript{1}, interlocutor P\textsubscript{2}) and topic. Exp2 introduces a topic unfamiliar to the target speaker while keeping the original pairing. In Exp3, we replace the original PRODIGy Speaker 2 with a character from a different movie (target P\textsubscript{1}, interlocutor P\textsubscript{rand}), while keeping the topic familiar with the target speaker; Exp4 randomises both Speaker 2 and the topic. 
For experiments involving non-PRODIGy speakers, Exp5 pairs the original PRODIGy Speaker 1 with a new non-PRODIGy Speaker 2 (target P\textsubscript{1}, interlocutor N\textsubscript{rand}) while maintaining the topic, whereas Exp6 uses both a random non-PRODIGy Speaker 2 and a random topic. Finally, Exp7 features an entirely new pairing of non-PRODIGy speakers (target N\textsubscript{1}, interlocutor N\textsubscript{2}) with a random topic. 

\begin{table}[ht!]
\centering
\small
\begin{tabular}{lllll}
\toprule
\textbf{Exp}       & \textbf{Target}   & \textbf{Interlocutor}      & \textbf{Topic} & \textbf{Pairing}\\ 
\midrule

Exp1   & P$_{1}$   & P$_{2}$      & Familiar & Familiar \\
Exp2   & P$_{1}$   & P$_{2}$      & Unfamiliar  & Familiar\\
Exp3   & P$_{1}$   & P$_{rand}$ & Familiar & Unfamiliar \\
Exp4   & P$_{1}$   & P$_{rand}$ & Unfamiliar & Unfamiliar\\
Exp5   & P$_{1}$   & N$_{rand}$ & Familiar & Unfamiliar\\
Exp6   & P$_{1}$   & N$_{rand}$ & Unfamiliar & Unfamiliar \\
Exp7   & N$_{1}$   & N$_{2}$      & Unfamiliar & Familiar \\
\bottomrule
\end{tabular}
\caption{
Overview of experimental configurations assessing the impact of speaker pairing and topic variation on dialogue generation.
}
\label{tab:experiments_overview}
\end{table}

\section{Detailed Automatic Results}\label{detailed-automatic-results}

\subsection{Fine-Tuned Model's Results}

In this section, we explore how the LLM-as-a-judge performs under various disclosure scenarios when identifying speakers in dialogues generated by the fine-tuned model, by systematically varying the information provided—ranging from full access to interlocutors’ biographies and dialogue turns to masking this information. The results, as outlined in Table \ref{tab:ft-classifier-res}, highlight the model’s adaptability and its limitations when confronted with unfamiliar or artificially constructed scenarios.

Revealing both the interlocutor’s biography and their previous turns yields high accuracy scores, 0.83 in Exp1 and 0.807 in Exp2. Providing only the biography while masking the turns still yields strong accuracy, 0.825 in Exp1 and 0.805 in Exp2. However, when both the biography and turns are masked, accuracy drops markedly to 0.587 in Exp1 and 0.570 in Exp2. These results suggest that the fine-tuned model effectively incorporates character biographies, even across varying topics, allowing the judge LLM to reliably identify speakers. 

The model's performance shifts significantly when we introduce more artificial scenarios, i.e., PRODIGy speakers coming from different movies (Exp3) or combinations of PRODIGy and Non-PRODIGy characters (Exp4). When both biography and dialogue turns are disclosed, accuracy drops noticeably to 0.558 in Exp3 and 0.560 in Exp4, substantially lower than in the original character scenarios. Interestingly, revealing only the interlocutor’s turns while masking the biography yields slightly improved scores (accuracy of 0.578 in Exp3 and 0.572 in Exp4). Conversely, providing only the biography while masking the turns lowers performance, with accuracy scores of 0.553 in Exp3 and 0.558 in Exp4.

The performance drop is even more pronounced when PRODIGy characters are paired with Non-PRODIGy characters. In these cases, even full disclosure of both biography and turns yields weaker scores (0.407 in Exp5 and 0.405 in Exp6). Providing only the interlocutor's biography while masking their turns results in slightly improved performance for Exp5 (0.414 accuracy) but lower scores for Exp6 (0.399 accuracy). Surprisingly, the best scores occur when both the interlocutor's biography and turns are masked (accuracy of 0.435 for Exp5 and 0.460 for Exp6). Detailed breakdowns for Exp5 and Exp6 in Speaker1 and Speaker2 metrics, as shown in Table~\ref{tab:exp-6-7-speakers-table-ft}, indicate these poor scores are due to the fact that the model struggles to effectively recognise non-PRODIGy interlocutors.

In the case of dialogues between unfamiliar non-PRODIGy speakers (Exp7), the highest accuracy (0.606) appears when only the biography is disclosed and the turns are masked. Providing both the biography and turns reduces performance slightly to 0.591. Masking the biography, however, causes a severe drop, with accuracy scores of 0.414 when only turns are disclosed and 0.415 when both biography and turns are masked.

\begin{table}[ht!]
\small
\centering
\begin{tabular}{llrrrr}
\toprule
\textbf{Exp}                   & \textbf{Disclosure} & \textbf{Acc} & \textbf{Prec} & \textbf{Rec} & \textbf{F1} \\ \midrule

\multirow{4}{*}{\textbf{Exp1}} & Both$_{Disc}$             & \textbf{0.830}         & \textbf{0.846}         & \textbf{0.825}        & \textbf{0.828}       \\
                               & Bio$_{Disc}$             & 0.825        & 0.841         & 0.820         & 0.822       \\
                               & Turns$_{Disc}$             & 0.601        & 0.640          & 0.596        & 0.592       \\
                               & Both$_{Mask}$             & 0.587        & 0.620          & 0.582        & 0.577       \\ \midrule
\multirow{4}{*}{\textbf{Exp2}} & Both$_{Disc}$             & \textbf{0.807}        & \textbf{0.825}         & \textbf{0.808}        & \textbf{0.808}       \\
                               & Bio$_{Disc}$             & 0.805        & 0.823         & 0.806        & 0.804       \\
                               & Turns$_{Disc}$             & 0.565        & 0.604         & 0.566        & 0.561       \\
                               & Both$_{Mask}$             & 0.570         & 0.606         & 0.571        & 0.567       \\ \midrule
\multirow{4}{*}{\textbf{Exp3}} & Both$_{Disc}$             & 0.558        & 0.591         & 0.561        & 0.554       \\
                               & Bio$_{Disc}$             & 0.553        & 0.578         & 0.556        & 0.549       \\
                               & Turns$_{Disc}$             & \textbf{0.578}        & \textbf{0.622}         & \textbf{0.582}        & \textbf{0.573}       \\
                               & Both$_{Mask}$             & 0.565        & 0.604         & 0.568        & 0.560        \\ \midrule
\multirow{4}{*}{\textbf{Exp4}} & Both$_{Disc}$             & 0.560        & 0.582         & 0.558        & 0.551       \\
                               & Bio$_{Disc}$             & 0.558        & 0.578         & 0.556        & 0.549       \\
                               & Turns$_{Disc}$             & \textbf{0.572}        & \textbf{0.609}         & \textbf{0.570}         & 0.561       \\
                               & Both$_{Mask}$             & \textbf{0.572}        & 0.607         & \textbf{0.570}         & \textbf{0.562}       \\ \midrule
\multirow{4}{*}{\textbf{Exp5}} & Both$_{Disc}$             & 0.407        & 0.429         & 0.409        & 0.398       \\
                               & Bio$_{Disc}$             & 0.414        & 0.436         & 0.416        & 0.403       \\
                               & Turns$_{Disc}$             & 0.434        & \textbf{0.469}         & 0.436        & 0.421       \\
                               & Both$_{Mask}$             & \textbf{0.435}        & 0.465         & \textbf{0.438}        & \textbf{0.423}       \\ \midrule
\multirow{4}{*}{\textbf{Exp6}} & Both$_{Disc}$             & 0.405        & 0.421         & 0.411        & 0.396       \\
                               & Bio$_{Disc}$             & 0.399        & 0.414         & 0.406        & 0.391       \\
                               & Turns$_{Disc}$             & 0.458        & 0.490          & 0.466        & 0.450        \\
                               & Both$_{Mask}$             & \textbf{0.460}         & \textbf{0.491}         & \textbf{0.467}        & \textbf{0.452}       \\ \midrule
\multirow{4}{*}{\textbf{Exp7}} & Both$_{Disc}$             & 0.591        & 0.631         & 0.589        & 0.584       \\
                               & Bio$_{Disc}$             & \textbf{0.606}        & \textbf{0.637}         & \textbf{0.605}        & \textbf{0.602}       \\
                               & Turns$_{Disc}$             & 0.414        & 0.433         & 0.412        & 0.397       \\
                               & Both$_{Mask}$             & 0.415        & 0.431         & 0.412        & 0.395       \\ \bottomrule
\end{tabular}
\caption{Judge LLM results fine-tuned model's generations}
\label{tab:ft-classifier-res}
\end{table}

\begin{table*}[ht!]
\small
\centering
\begin{tabular}{llrr|rr|rr|rr}\toprule
                      &       & \multicolumn{2}{c}{\textbf{Acc}} & \multicolumn{2}{c}{\textbf{Prec}} & \multicolumn{2}{c}{\textbf{Rec}} & \multicolumn{2}{c}{\textbf{F1}} \\ \toprule
                      & \textbf{Disclosure} & \textbf{Sp1}        & \textbf{Sp2}        & \textbf{Sp1}         & \textbf{Sp2}        & \textbf{Sp1}        & \textbf{Sp2}        & \textbf{Sp1}        & \textbf{Sp2}       \\ \midrule
\multirow{4}{*}{\textbf{Exp5}} & Both$_{Disc}$    & 0.552      & 0.262      & 0.578       & 0.270       & 0.559      & 0.261      & 0.545      & 0.248     \\
                      & Bio$_{Disc}$    & 0.541      & 0.286      & 0.571       & 0.292      & 0.548      & 0.284      & 0.533      & 0.270      \\
                      & Turns$_{Disc}$    & \textbf{0.562}      & 0.306      & \textbf{0.600}         & 0.323      & 0.569      & 0.304      & 0.554      & 0.284     \\
                      & Both$_{Mask}$    & 0.565      & 0.306      & 0.599       & 0.313      & \textbf{0.572}      & 0.303      & \textbf{0.558}      & 0.283     \\ \midrule
\multirow{4}{*}{\textbf{Exp6}} & Both$_{Disc}$    & 0.530       & 0.280      & 0.548       & 0.287      & 0.534      & 0.290       & 0.523      & 0.269     \\
                      & Bio$_{Disc}$    & 0.515      & 0.283      & 0.539       & 0.286      & 0.519      & 0.292      & 0.509      & 0.273     \\
                      & Turns$_{Disc}$    & 0.563      & 0.354      & 0.592       & 0.379      & 0.567      & 0.364      & 0.556      & 0.341     \\
                      & Both$_{Mask}$    & \textbf{0.574}      & 0.346      & \textbf{0.614}       & 0.364      & \textbf{0.580}       & 0.353      & \textbf{0.569}      & 0.335    \\ \bottomrule
\end{tabular}
\caption{Judge LLM results for Exp5 and Exp6 for Speaker 1 and Speaker 2 identification in fine-tuned model's generations.}
\label{tab:exp-6-7-speakers-table-ft}
\end{table*}

\subsection{Zero-Shot Results}

Table~\ref{tab:noZS-classifier-res} presents the metrics from our zero-shot experiments. Compared to the fine-tuned model's results, these zero-shot evaluations follow similar trends but consistently yield higher scores.

When dealing with dialogues generated using original PRODIGy speaker pairings (Exp1 and Exp2), the judge model performs particularly well when it has access to the interlocutor’s biography. In Exp1 (with original topic), disclosing both biography and turns yields an accuracy of 0.882, while disclosing the biography alone still achieves 0.878. Similarly, in Exp2 (random topic), accuracy remains high at 0.887 and 0.888, respectively. 
In contrast, masking the biography leads to a performance drop. When only the interlocutor’s turns are disclosed, accuracy declines to 0.750 in Exp1 and 0.774 in Exp2, and further drops when both biography and turns are masked, reaching 0.737 in Exp1 and 0.730 in Exp2.

When PRODIGy speakers come from different movies or interact with Non-PRODIGy characters, performance declines. In these scenarios (Exp3 and Exp4), biographical information offers no clear advantage, while focusing on interlocutor turns or masking all interlocutor details proves more effective. In Exp3, disclosing only turns yields an accuracy of 0.718, slightly improving to 0.722 when both biography and turns are masked. In contrast, revealing both interlocutor's biography and turns lowers accuracy to 0.701, and providing only the biography results in an even lower 0.694. A similar pattern emerges in Exp4, where disclosing only turns achieves the highest accuracy (0.725), followed by masking both (0.712), while approaches that reveal biography perform worse (0.691 and 0.676, respectively). 

Performance degradation is even more pronounced when pairing PRODIGy characters with Non-PRODIGy characters (Exp5 and Exp6). In these cases, disclosing interlocutor's biographical information actually hinders performance compared to relying solely on interlocutor's turns or masking both elements. In Exp5, the highest accuracy (0.730) is achieved by masking both interlocutor's biography and turns, followed closely by disclosing only their turns (0.723). In contrast, providing interlocutor's biography leads to significantly worse performance, with accuracy dropping to 0.657 when both biography and turns are disclosed and 0.654 when only biography is revealed. Exp6 follows a similar trend, where the best results occur when both elements are masked (0.731) or when only turns are disclosed (0.723), while disclosing biography results in the lowest scores (0.645 with both biography and turns disclosed, 0.636 with only biography disclosed).

However, when evaluating dialogues between non-PRODIGy speakers (Exp7), performance are similar to those observed in the original PRODIGy pairing scenarios. The highest accuracy score (0.826) is achieved when only the interlocutor's biography is disclosed, with nearly identical performance (0.824 accuracy) when both biography and turns are provided. Performance declines when biographical information is masked, dropping to accuracy scores of 0.778 when only turns are disclosed and 0.786 when both turns and biography are masked.

\begin{table}[ht!]
\small
\centering
\begin{tabular}{llrrrr}\toprule
\textbf{Exp}          & \textbf{Disclosure} & \textbf{Acc}   & \textbf{Prec}  & \textbf{Rec}   & \textbf{F1}    \\ \toprule
\multirow{4}{*}{\textbf{Exp1}} & Both$_{Disc}$             & \textbf{0.882} & \textbf{0.889} & \textbf{0.883} & \textbf{0.882} \\
                      & Bio$_{Disc}$             & 0.878          & 0.884          & 0.879          & 0.878          \\
                      & Turns$_{Disc}$             & 0.750           & 0.777          & 0.751          & 0.749          \\
                      & Both$_{Mask}$             & 0.737          & 0.755          & 0.738          & 0.736          \\ \midrule
\multirow{4}{*}{\textbf{Exp2}} & Both$_{Disc}$             & 0.887          & \textbf{0.898} & 0.886          & \textbf{0.888} \\
                      & Bio$_{Disc}$             & \textbf{0.888} & \textbf{0.898} & \textbf{0.887} & \textbf{0.888} \\
                      & Turns$_{Disc}$             & 0.774          & 0.799          & 0.772          & 0.773          \\
                      & Both$_{Mask}$             & 0.730           & 0.748          & 0.728          & 0.728          \\ \midrule
\multirow{4}{*}{\textbf{Exp3}} & Both$_{Disc}$             & 0.701          & 0.718          & 0.700            & 0.698          \\
                      & Bio$_{Disc}$             & 0.694          & 0.707          & 0.693          & 0.692          \\
                      & Turns$_{Disc}$             & 0.718          & \textbf{0.749} & 0.717          & 0.717          \\
                      & Both$_{Mask}$             & \textbf{0.722} & 0.745          & \textbf{0.721} & \textbf{0.720}  \\ \midrule
\multirow{4}{*}{\textbf{Exp4}} & Both$_{Disc}$             & 0.691          & 0.705          & 0.692          & 0.690           \\
                      & Bio$_{Disc}$             & 0.676          & 0.689          & 0.677          & 0.673          \\
                      & Turns$_{Disc}$             & \textbf{0.725} & \textbf{0.751} & \textbf{0.726} & 0.723          \\
                      & Both$_{Mask}$             & 0.712          & 0.735          & 0.713          & \textbf{0.710}  \\ \midrule
\multirow{4}{*}{\textbf{Exp5}} & Both$_{Disc}$             & 0.657          & 0.672          & 0.652          & 0.653          \\
                      & Bio$_{Disc}$             & 0.654          & 0.667          & 0.649          & 0.650           \\
                      & Turns$_{Disc}$             & 0.723          & \textbf{0.752} & 0.718          & 0.722          \\
                      & Both$_{Mask}$             & \textbf{0.730}  & 0.749          & \textbf{0.725} & \textbf{0.727} \\ \midrule
\multirow{4}{*}{\textbf{Exp6}} & Both$_{Disc}$             & 0.645          & 0.665          & 0.644          & 0.642          \\
                      & Bio$_{Disc}$             & 0.638          & 0.659          & 0.638          & 0.636          \\
                      & Turns$_{Disc}$             & 0.723          & 0.759          & 0.722          & 0.722          \\
                      & Both$_{Mask}$             & \textbf{0.731} & \textbf{0.761} & \textbf{0.731} & \textbf{0.730}  \\ \midrule
\multirow{4}{*}{\textbf{Exp7}} & Both$_{Disc}$             & 0.824          & 0.839          & 0.821          & 0.822          \\
                      & Bio$_{Disc}$             & \textbf{0.826} & \textbf{0.840}  & \textbf{0.823} & \textbf{0.824} \\
                      & Turns$_{Disc}$             & 0.778          & 0.807          & 0.773          & 0.774          \\
                      & Both$_{Mask}$             & 0.786          & 0.809          & 0.781          & 0.783         \\ \bottomrule
\end{tabular}
\caption{Judge LLM results on zero-shot generations}
\label{tab:noZS-classifier-res}
\end{table}

\subsection{Overlap Metrics and Rare Words}\label{overlap-metrics}

To assess how extensively models replicate information from the target biographies into generated target turns, 
we examined the incorporation of specific rare words from the target biographies into the generated speaker turns. When a specific rare word from a biography appears in the turns, it enhances the recognisability of the target speaker. This pattern indicates a tendency for the model to copy and paste biographical information into the dialogue.

For this analysis, we employed the \texttt{wordfreq} library\footnote{\url{https://github.com/rspeer/wordfreq/}}, utilising its \texttt{zipf\_frequency} function, which ranks word rarity on a logarithmic scale. This scale represents the base-10 logarithm of the frequency per billion words \cite{van2014subtlex}. For instance, a word with a Zipf value of 6 appears once per thousand words, while a value of 3 corresponds to once per million words. In our analysis, we defined rare words as those with a Zipf frequency below 4.0, focusing on infrequent terms to better capture copying tendencies.

Table~\ref{tab:rare-word-overlap} shows the percentage of dialogues where at least one rare word from the target speaker's biography also appears in their dialogue turns. Across all experiments, the zero-shot (ZS) setting consistently demonstrates higher overlap percentages than the fine-tuned (FT) model. For example, in Exp1, ZS reaches 41.84\% compared to FT's 9.68\%. The gold test set is included for comparison, showing a much lower overlap of 5.84\%. These results highlight the zero-shot models' tendency to replicate rare biographical terms in generated dialogues.

\begin{table}[ht!]
\small
\centering
\begin{tabular}{lrr}\toprule
              & \multicolumn{2}{l}{\textbf{\% dialogues}} \\ \midrule
\textbf{Test} & \multicolumn{2}{c}{5.84\%}       \\ \midrule
\textbf{}     & \textbf{FT}    & \textbf{ZS}    \\ \midrule
\textbf{Exp1} & 9.68\%         & 41.84\%         \\
\textbf{Exp2} & 7.84\%         & 39.76\%         \\
\textbf{Exp3} & 5.04\%         & 42.64\%         \\
\textbf{Exp4} & 6.24\%         & 40.64\%         \\
\textbf{Exp5} & 3.28\%         & 40.00\%         \\
\textbf{Exp6} & 3.36\%         & 36.48\%         \\
\textbf{Exp7} & 0.72\%         & 43.44\%    \\\bottomrule    
\end{tabular}
\caption{Percentage of dialogues where the target speaker's biography and dialogue turns share at least one rare word (threshold 4.0), comparing Fine-Tuned (FT) and Zero-Shot (ZS) settings.
}
\label{tab:rare-word-overlap}
\end{table}

Besides the use of rare words, we also computed overlap scores using METEOR, BLEU-1, and ROUGE-1 metrics\footnote{To compute the metrics we employed the Evaluate library (\url{https://huggingface.co/docs/evaluate/}).}. The results, detailed in Table~\ref{tab:overlap-metrics}, include comparisons with the overlap scores between biographies and the gold turns. The results reveal that overlap metrics are significantly higher in the zero-shot setting compared to the fine-tuned setting across all configurations (e.g., Exp1 METEOR: FT 0.047 vs ZS 0.131; Exp7 METEOR: FT 0.036 vs ZS 0.153). Additionally, the scores for the fine-tuned model more closely match those of the gold dialogues (e.g., METEOR 0.032), suggesting an adaptation to the dialogues' gold characteristics.

\begin{table}[ht!]
\small
\centering
\begin{tabular}{lrrrrrr} \toprule
              & \multicolumn{2}{c}{\textbf{METEOR}} & \multicolumn{2}{c}{\textbf{BLEU-1}} & \multicolumn{2}{c}{\textbf{ROUGE-1}} \\  \toprule
              
\textbf{Test} & \multicolumn{2}{c}{0.032}           & \multicolumn{2}{c}{0.016}           & \multicolumn{2}{c}{0.051}            \\ \midrule
& \textbf{FT}      & \textbf{ZS}     & \textbf{FT}      & \textbf{ZS}     & \textbf{FT}      & \textbf{ZS}      \\ \midrule
\textbf{Exp1} & 0.047            & 0.131            & 0.043            & 0.155            & 0.045            & 0.123             \\
\textbf{Exp2} & 0.046            & 0.129            & 0.043            & 0.154            & 0.044            & 0.123             \\
\textbf{Exp3} & 0.044            & 0.142            & 0.041            & 0.158            & 0.042            & 0.133             \\
\textbf{Exp4} & 0.045            & 0.138            & 0.041            & 0.156            & 0.043            & 0.130              \\
\textbf{Exp5} & 0.042            & 0.144            & 0.036            & 0.171            & 0.041            & 0.147             \\
\textbf{Exp6} & 0.042            & 0.141            & 0.036            & 0.170             & 0.042            & 0.145             \\
\textbf{Exp7} & 0.036            & 0.153            & 0.027            & 0.182            & 0.039            & 0.164            \\ \bottomrule
\end{tabular}
\caption{Overlap metrics between speakers' biographies and turns: fine-tuned model's generations (FT) vs zero-shot generations (ZS)}
\label{tab:overlap-metrics}
\end{table}

\section{Human Evaluation Guidelines}\label{human-eval-appendix}

Table~\ref{tab:human-guidelines} outlines the instructions provided to human evaluators for the evaluation task. Evaluators are presented with dialogues between two speakers and must identify the correct biography (Profile A, Profile B, or Profile C) for one of the speakers. 

\humanevalguidelines{}{
You will be presented with a series of dialogues between two speakers.

For each dialogue your task is to guess the correct biography (Profile) of one of the two speakers.

There are 3 possible Profiles to choose among (shown at the end of the dialogue as "Profile A", "Profile B", or "Profile C"). Please make your guess even though the dialogue may sound a little weird or unnatural. You will also have the possibility to leave any comment you deem relevant about each dialogue/task in the “comments” field.\\\\

\textbf{IMPORTANT}: for each dialogue we have different configurations where we either disclose or hide some of the interlocutor’s information:\\\\

- We will either provide or not interlocutor profile, in the latter case we use the tag [MASKED] \\\\
- We will either show or mask other speakers’ turns in the dialogue.}{Human Evaluation Guidelines}{human-guidelines}

\section{Fine-Tuned Model's Generation Examples}
In this section, we showcase four dialogues generated by the fine-tuned model. The first dialogue, involving an original PRODIGy speaker pairing (James Cole and Dr. Railly from the \textit{12 Monkeys} movie), accurately captures Cole’s paranoia and Railly’s composed professionalism. The second dialogue pairs PRODIGy characters from different movies (Walter Sobchak from \textit{The Big Lebowski} and Spider-Man), effectively preserving Walter’s confrontational nature and Spider-Man’s guilt-ridden introspection. The third dialogue involves a PRODIGy character (again Walter) interacting with a Non-PRODIGy character (a data analyst), showcasing Walter’s strong opinions against the analyst’s more reserved responses. The fourth dialogue, between two Non-PRODIGy characters (a high school student and a psychologist), maintains an introspective tone, reflecting their emotional depth. Across all dialogues, the model successfully aligns responses with the given persona traits, ensuring consistency.

\dialoguetable{1}{\textbf{Prompt:} \\\\ 
\noindent You are \textbf{Speaker1}. Your biography is as follows:\\\\

\begin{itemize}[leftmargin=*, itemsep=2pt, parsep=0pt]
    \item I live in what is left of Philadelphia in 2035.
    \item I am single, but I am developing feelings for my psychiatrist, Dr. Railly.
    \item I am a total nutcase, completely deranged, delusional, paranoid.
    \item I am visiting 1990, after being tossed in jail; I have been sent back in time to learn about a virus before it started mutating.
    \item When I told others I was from the future, they thought I was out of my mind and threw me in an insane asylum.
\end{itemize}

\noindent You are having a dialogue with \textbf{Speaker2}. Their biography is as follows:
\begin{itemize}[leftmargin=*, itemsep=2pt, parsep=0pt]
    \item I am an intelligent and even-keeled psychiatrist.
    \item Philadelphia police asked me to diagnose the sanity of James Cole, a prisoner living in an underground compound.
    \item I am a good-natured, idealistic woman.
    \item I am arrogant and self-assured, but not obnoxiously smug like some of my colleagues.
    \item I am helpful and kind, and genuinely care about my patients.
\end{itemize}

\textbf{Dialogue:} \\\\

\noindent\textbf{Speaker1:} Dr. Railly... please help him! He's going to kill us all -- 

\noindent\textbf{Speaker2:} James... you need to calm down now. You can't let yourself get this upset because if we're going to work together here you both have to be able to stay focused on the task at hand which is identifying who he might be and why he did this to himself. Okay?

\noindent\textbf{Speaker1:} Who?!

\noindent\textbf{Speaker2:} His name is probably someone he trusts or has access too.

\noindent\textbf{Speaker1:} Okay okay -- I know that already -- his wife maybe...

\noindent\textbf{Speaker2:} Let's talk to each other very clearly right now. What do we know about Joseph Dunn?

\noindent\textbf{Speaker1:} He killed seven people then turned the gun on himself... Oh wait, no one knows who he really is yet... so there's nothing to say.

\noindent\textbf{Speaker2:} Exactly. And now let's go over everything else that happened last night when these things were occurring.

}{A dialogue between James Cole and Dr. Cassandra Railly from the movie \textit{12 Monkeys}}{exp-1-dialogue}

\dialoguetable{2}{\textbf{Prompt:} \\\\ 
\noindent You are \textbf{Speaker1}. Your biography is as follows:\\\\

\begin{itemize}[leftmargin=*, itemsep=2pt, parsep=0pt]
    \item I am owner of Sobchak Security; I use the combat knowledge I accumulated during the Vietnam War.
    \item I am divorced from my ex-wife Cynthia after five years of marriage.
    \item I live in Southern California.
    \item I spend most of my time at the local bowling alley and picking fights with those who don't respect the rules.
    \item I am stubborn, aggressive, and a mentally unhinged believer in rules.
\end{itemize}

\noindent You are having a dialogue with \textbf{Speaker2}. Their biography is as follows:\\

\begin{itemize}[leftmargin=*, itemsep=2pt, parsep=0pt]
    \item I am a superhero and freelance photographer taking and selling pictures of myself as Spider-Man.
    \item I am kind, reserved, clever, a lifelong outcast, and I tend to be quiet.
    \item Whenever I am Spider-Man, I become a consummate jokester, using one-liners to irritate my foes.
    \item I am living an unrequited love: I have always had a crush on my neighbor Mary Jane Watson, who seems out of my reach.
    \item I am super strong, super agile, capable of sensing danger, and able to shoot webs out of my hands.
\end{itemize}

\textbf{Dialogue:} \\\\

\noindent\textbf{Speaker1:} He's coming home! He's coming home! You gotta get back up there!

\noindent\textbf{Speaker2:} How can you expect me --

\noindent\textbf{Speaker1:} It doesn't matter how you got here. What matters is that it was your job to save him now that he needs help. And if you walk away without doing something for this guy, then you're not Superman... you're just a bunch of hot air!!

\noindent\textbf{Speaker2:} That's enough!!!

\noindent\textbf{Speaker1:} No -- it ain't! Not when someone else has done what you couldn't do. Not while I've stood by like some damn cheerleader. It ends tonight, Spidey -- and so does our partnership.

\noindent\textbf{Speaker2:} Wait... wait before you go, tell them everything. Please. Let me explain. Everything happened too fast. My father was dying. That cop died because of me. They thought they could kill another man off like a spider -- like "Spider-Man." And I did nothing until I found MJ. She needed help.

\noindent\textbf{Speaker1:} She? The girl? Yes, but she didn't need all the rest of us either -- or anything else except you. Now listen to me: we gonna take care of Mr. Parker. We'll find his friends. We'll make some calls, see if anyone wants to buy these goddamn pictures.

\noindent\textbf{Speaker2:} What?!
}{A dialogue between Walter Sobchak from \textit{The Big Lebowski} and Spider-Man}{exp-2-dialogue}

\dialoguetable{3}{\textbf{Prompt:} \\\\ 
\noindent You are \textbf{Speaker1}. Your biography is as follows:\\\\

\begin{itemize}[leftmargin=*, itemsep=2pt, parsep=0pt]
    \item I am owner of Sobchak Security; I use the combat knowledge I accumulated during the Vietnam War.
    \item I am divorced from my ex-wife Cynthia after five years of marriage.
    \item I live in Southern California.
    \item I spend most of my time at the local bowling alley and picking fights with those who don't respect the rules.
    \item I am stubborn, aggressive, and a mentally unhinged believer in rules.
\end{itemize}

\noindent You are having a dialogue with \textbf{Speaker2}. Their biography is as follows:\\\\

\begin{itemize}[leftmargin=*, itemsep=2pt, parsep=0pt]
    \item I work as a data analyst in a tech company, where I enjoy solving complex problems with numbers.
    \item As an introvert, I often find solace in my own company and prefer small gatherings to large parties.
    \item I have a deep love for Marvel movies and comics, and I get really anxious when I encounter spoilers.
    \item In my free time, I enjoy reading graphic novels and binge-watching the latest superhero series on streaming platforms.
    \item I live alone in a cozy apartment filled with Marvel memorabilia and cozy blankets.
\end{itemize}

\textbf{Dialogue:} \\\\

\noindent\textbf{Speaker1:} Do you know what this place used to be?

\noindent\textbf{Speaker2:} The Satchel Pig restaurant... What's it now? A movie set?

\noindent\textbf{Speaker1:} A motel. It was called the El Rancho Motel. The best damn motel between LA and San Francisco.

\noindent\textbf{Speaker2:} Well, that explains why they didn't tear it down.

\noindent\textbf{Speaker1:} People around here, they've got too much loyalty. They're like family. You can't just knock them out. That's one thing wrong with America today, we lack our sense of community. We all need more loyalty. Like me.

\noindent\textbf{Speaker2:} Yeah right.

\noindent\textbf{Speaker1:} They used to hold parties there every Friday night. All kinds of people would come—businessmen, prostitutes, truckers, you name it. And there'd always be music, dancing, girls doing their hair under strobe lights. They were wild times. This building over here, it's the old lounge bar. The bartender still works there—Richie. He remembers me.

\noindent\textbf{Speaker2:} What did he say?
}{A dialogue between Walter Sobchak from \textit{The Big Lebowski} and a Non-PRODIGy character}{exp-6-dialogue}

\dialoguetable{4}{\textbf{Prompt:} \\\\ 
\noindent You are \textbf{Speaker1}. Your biography is as follows:\\\\

\begin{itemize}[leftmargin=*, itemsep=2pt, parsep=0pt]
    \item I am a high school student who finds inspiration in literature, often losing myself in the worlds created by authors.
    \item Currently, I work part-time at a local bookstore, where I enjoy helping customers find their next great read.
    \item In my free time, I volunteer at a community center, where I assist with after-school programs for younger kids.
    \item I value my close relationships, but I tend to prefer smaller gatherings over large parties, as I feel more comfortable in intimate settings.
    \item My family is supportive of my goals, and they encourage me to pursue my interests in both reading and service.
\end{itemize}

\noindent You are having a dialogue with \textbf{Speaker2}. Their biography is as follows:\\\\

\begin{itemize}[leftmargin=*, itemsep=2pt, parsep=0pt]
    \item I am a clinical psychologist specializing in cognitive behavioral therapy, primarily working with clients who struggle with mood disorders.
    \item My passion for understanding the human mind stemmed from a personal journey through anxiety and depression during my teenage years.
    \item I enjoy spending my evenings reading about the latest research in psychology or delving into a good novel, often in the comfort of my home.
    \item In my professional life, I strive to create a warm and supportive environment for my clients, helping them navigate their challenges.
    \item I am currently single, focusing on my career and personal growth, though I do cherish the close friendships I have built over the years.
\end{itemize}

\textbf{Dialogue:} \\\\

\noindent\textbf{Speaker1:} Gosh -- thanks! You know that's really sweet of you.

\noindent\textbf{Speaker2:} Well -- I think it was kind of helpful for me too --

\noindent\textbf{Speaker1:} Really? How so?

\noindent\textbf{Speaker2:} It made me realize how much I'd like to be able to help people somehow, which has been something I've always wanted to do, ever since college when I first started volunteering here.

\noindent\textbf{Speaker1:} That makes sense... because this place seems pretty cool to hang out at anyway.

\noindent\textbf{Speaker2:} Yeah well -- yeah -- there isn't exactly a lot going on around here, if you can imagine --

\noindent\textbf{Speaker1:} But maybe being stuck here is better than most places -- you're making such progress. That would make any town seem fun. Are we doing okay here? Can I try asking another question? Because what I want to ask is why your face looks sad all the time. Like underneath everything else -- even when things look fine up top, deep down it just feels wrong. Is that true?

\noindent\textbf{Speaker2:} Yes. But let me explain before you get upset... It's not anything specific. More just knowing bad stuff happens to everyone sooner or later no matter how lucky or happy they appear otherwise. Anyway, getting back to you -- don't worry. We'll keep hanging out -- one way or another. All set on that end.
}{A dialogue between a high school student and a clinical psychologist}{exp-7-dialogue}

\end{document}